\documentclass{article}

\PassOptionsToPackage{authoryear, round}{natbib}

\usepackage[preprint]{neurips_2025}

\usepackage[utf8]{inputenc} 
\usepackage[T1]{fontenc}    
\usepackage{hyperref}       
\usepackage{url}            
\usepackage{booktabs}       
\usepackage{amsfonts}       
\usepackage{nicefrac}       
\usepackage{microtype}      
\usepackage{xcolor}         
\usepackage[most]{tcolorbox}
\usepackage{enumitem}
\usepackage{listings}
\usepackage{fancyvrb}
\usepackage{caption}
\usepackage{amsmath}
\usepackage{multirow}
\usepackage{graphicx}
\usepackage[capitalise, noabbrev]{cleveref}
\usepackage{textcomp}
\usepackage{wrapfig}

\title{CReSt: A Comprehensive Benchmark for Retrieval-Augmented Generation with Complex Reasoning over Structured Documents}

\author{%
  Minsoo Khang\thanks{Equal contribution.} \\
  Upstage AI \\
  \texttt{mkhang@upstage.ai} \\
  \And
  Sangjun Park\footnotemark[1] \\
  Upstage AI \\
  \texttt{sangjun@upstage.ai} \\
  \And
  Teakgyu Hong\\
  Upstage AI \\
  \texttt{teakgyu.hong@upstage.ai} \\
  \And
  Dawoon Jung\thanks{Corresponding author.} \\
  Upstage AI \\
  \texttt{dawoon@upstage.ai} \\
}

\begin{document}

\maketitle

\begin{abstract}
Large Language Models (LLMs) have made substantial progress in recent years, yet evaluating their capabilities in practical Retrieval-Augmented Generation (RAG) scenarios remains challenging. In practical applications, LLMs must demonstrate complex reasoning, refuse to answer appropriately, provide precise citations, and effectively understand document layout. These capabilities are crucial for advanced task handling, uncertainty awareness, maintaining reliability, and structural understanding. While some of the prior works address these aspects individually, there is a need for a unified framework that evaluates them collectively in practical RAG scenarios. To address this, we present CReSt (A Comprehensive Benchmark for Retrieval-Augmented Generation with \textbf{C}omplex \textbf{Re}asoning over \textbf{St}ructured Documents), a benchmark designed to assess these key dimensions holistically. CReSt comprises 2,245 human-annotated examples in English and Korean, designed to capture practical RAG scenarios that require complex reasoning over structured documents.
It also introduces a tailored evaluation methodology to comprehensively assess model performance in these critical areas. Our evaluation shows that even advanced LLMs struggle to perform consistently across these dimensions, underscoring key areas for improvement. We release CReSt to support further research and the development of more robust RAG systems. The dataset and code are available at: \url{https://github.com/UpstageAI/CReSt}.
\end{abstract}

\section{Introduction}
Large Language Models (LLMs) have demonstrated remarkable capabilities across a wide range of applications \citep{gpt4, llama2024}, yet they still exhibit limitations. To address these shortcomings, Retrieval-Augmented Generation (RAG) has emerged as a key paradigm, enhancing LLM performance by grounding responses in external knowledge sources \citep{gao2023rag}.

RAG pipelines often rely on web pages or PDF documents as external sources of knowledge \citep{tan2025htmlrag, tanaka2025vdocrag} and these applications require complex reasoning over semi-structured documents \citep{fan2024survey}. To effectively serve such use-cases, LLMs must possess a combination of critical skills:
(1) the ability to perform complex reasoning,
(2) the ability to appropriately refuse to answer when provided information is insufficient,
(3) the ability to cite the supporting document as evidence, and 
(4) the ability to understand documents in structured format, such as HTML.

In this paper, we introduce a benchmark designed to holistically evaluate multiple capabilities of LLMs in document-based RAG scenarios.
CReSt is constructed entirely from scratch using realistic source documents, unlike prior benchmarks that rely on refining existing datasets or draw heavily from overlapping sources such as Wikipedia.
These documents are parsed into both plain-text and HTML formats based on which we generate question–answer (QA) pairs that reflect the complexity and reasoning demands of real-world applications.
For comprehensive evaluation, we also include refusal cases \citep{chen2024rgb}, where questions are unanswerable due to insufficient information. Additionally, each QA pair is annotated with explicit citations that identify the supporting documents, facilitating precise grounding and verifiability. To generate complex reasoning questions, we design a QA generation method that synthesizes challenging examples in both English and Korean. These questions are subsequently revised by human annotators for quality assurance. 

Through our evaluation of several state-of-the-art LLMs, we observe that many models struggle with the CReSt benchmark, particularly in deciding whether they should refuse or not. Furthermore, we demonstrate that incorporating recently proposed reasoning strategies leads to performance gains, underscoring CReSt’s sensitivity to reasoning proficiency.
We hope our work contributes to the development of advanced RAG applications.


\section{Related Works}

\begin{table}[h]
  \centering
    \scriptsize
    \caption{Comparison of Document-based RAG Benchmarks. Multi-Dimensional Reasoning: benchmark requires multiple-types of reasoning. Structured Input: documents are provided in forms other than plain-text. \textsuperscript{*} represents cases where condition differs for different subset of the benchmark.}
    \label{tab:rag-benchmark-comparison}
    \begin{tabular}{@{}lccccc@{}}
    \toprule
    \textbf{Benchmark} & \textbf{Multi-Dimensional} & \textbf{Structured} & \textbf{Citation} & \textbf{Refusal} & \textbf{Language} \\
     & \textbf{Reasoning} & \textbf{Input} & & & \textbf{Coverage} \\
    \midrule
    HotpotQA~\citep{yang-2018-hotpotqa} & Yes & No         & Yes & No      & English           \\
    FEVER~\citep{thorne-2018-fever} & No  & No         & Yes & Yes     & English           \\
    KILT~\citep{petroni-2021-kilt} & Yes & No         & Yes & No\textsuperscript{*}     & English           \\
    ALCE~\citep{gao-2023-alce} & No  & No         & Yes & No      & English           \\
    RGB~\citep{chen2024rgb} & Yes  & No         & No  & Yes     & English, Chinese  \\
    CRUD-RAG~\citep{lyu2024crudrag} & Yes & No         & No  & No      & Chinese           \\
    UDA~\citep{hui2024uda} & Yes & Yes  & No  & No      & English           \\
    FRAMES~\citep{krishna-2025-frames} & Yes & Yes  & No  & No      & English           \\
    \midrule
    \textbf{CReSt (Ours)} & \textbf{Yes} & \textbf{Yes} & \textbf{Yes} & \textbf{Yes} & \textbf{English, Korean} \\
    \bottomrule
    \end{tabular}
\end{table}

RAG has rapidly emerged as a crucial paradigm for addressing the inherent limitations of LLMs, such as hallucinations and insufficient grounding in factual knowledge \citep{gao2023rag}. While various benchmarks have been introduced to assess RAG systems, existing studies typically focus on limited subsets of criteria, without considering real-world complexities which demand holistic evaluation of RAG systems, including complex reasoning, answer refusal capability, citation accuracy, and the ability to process diverse document formats. Table~\ref{tab:rag-benchmark-comparison} shows the comparison overview of CReSt against other benchmarks across various criteria.

Most existing benchmarks, such as HotPotQA \citep{yang-2018-hotpotqa}, FEVER \citep{thorne-2018-fever}, and KILT \citep{petroni-2021-kilt} predominantly utilize plain-text documents. However, real-world applications often involve HTML documents, which encode structural and formatting information. Extending these benchmarks to include HTML-formatted documents would enhance their applicability in practical RAG scenarios.
Recent benchmarks, such as ALCE \citep{gao-2023-alce}, RGB \citep{chen2024rgb}, and FRAMES \citep{krishna-2025-frames}, have introduced tasks targeting different aspects of RAG evaluation. ALCE focuses on citation generation for long-form questions; RGB assesses robustness against noisy or misleading contexts; and FRAMES incorporates multiple reasoning dimensions including tabular, temporal, and numerical reasoning.

While these benchmarks address important aspects critical in RAG evaluations, they still fall short of providing a holistic evaluation framework for practical deployments. In contrast, our benchmark addresses these significant gaps by evaluating integrated capabilities essential for realistic, complex RAG applications. We incorporate both plain-text and HTML documents to better reflect real-world knowledge sources, systematically generate complex reasoning questions, mandate citation accuracy, and include explicit scenarios for refusal when valid answers are unavailable. By combining these elements, our work presents a more comprehensive and robust framework for evaluating advanced RAG systems, thereby contributing meaningfully towards their practical deployment.

\section{Dataset}
\begin{figure}[t]
\centering
\includegraphics[width=0.8\textwidth]{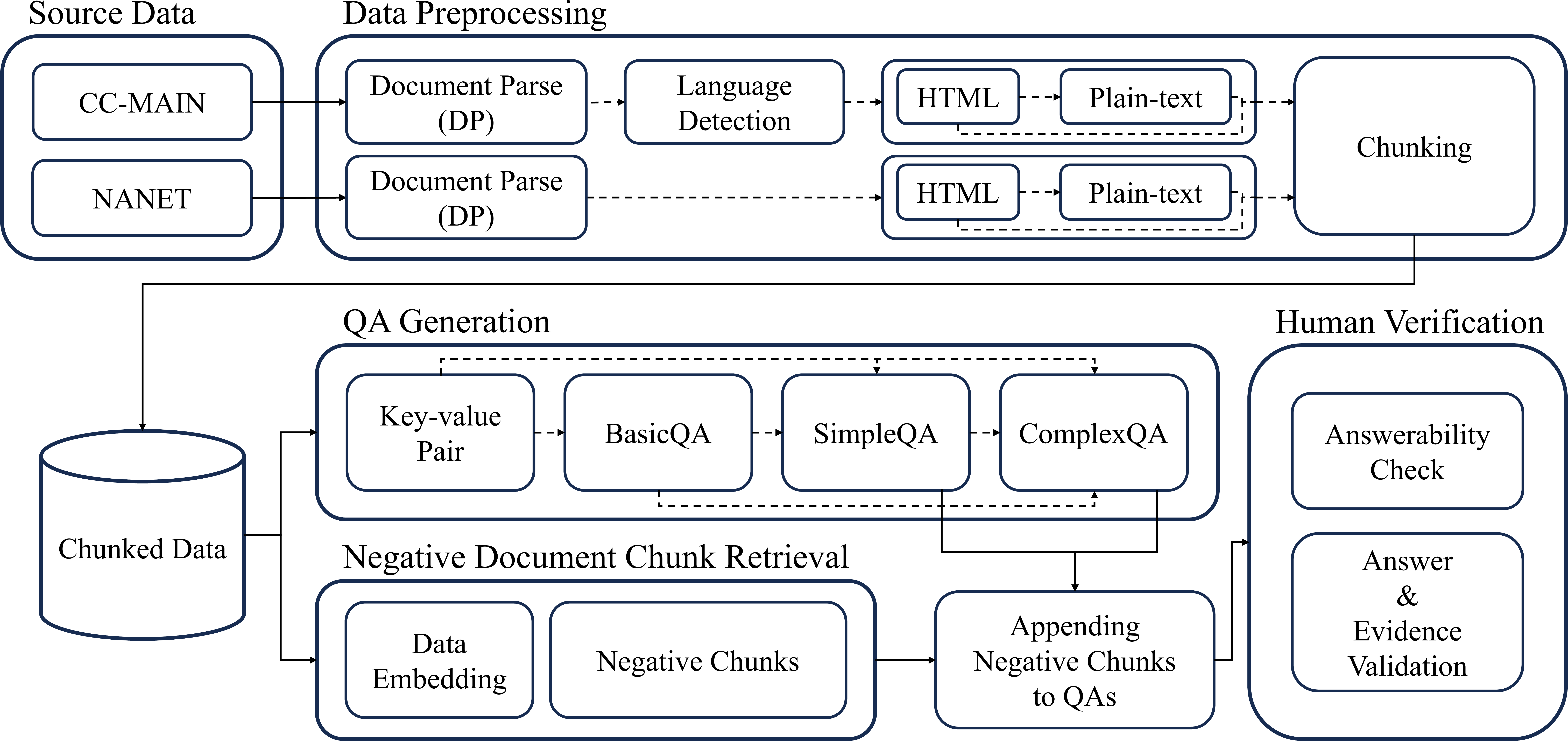}
\caption{Dataset construction pipeline.}
\label{fig:general_datagen_pipeline}
\end{figure}

CReSt is a RAG benchmark comprising over 2,000 data instances in both English and Korean. An overview of the dataset construction pipeline is provided in \cref{fig:general_datagen_pipeline}. Each instance is structured as: \textit{<Document chunk(s), Query, Answer, Citation indices>}. The document chunks include both plain-text and HTML formats and are categorized into positive and negative chunks. Positive chunks contain relevant evidence necessary to answer the query, while negative chunks represent similar topics but unnecessary information.
CReSt comprises over 20,000 document chunks, with a well-balanced distribution between plain-text and HTML representations ($\sim$55\% HTML).
The benchmark includes citation labels that identify the specific positive chunks associated with each question–answer pair, enabling precise referencing and verification of the evidence.

To ensure broad document-domain coverage in both English and Korean, CReSt sources raw documents from two publicly available collections: PDF files from Common Crawl (CC-MAIN) for English and crawled document images from National Assembly Library (NANET) of Korea for Korean.
From each source, 2,000 documents are randomly sampled to initiate the dataset construction process.
These are subsequently converted into both HTML and plain-text representations using a publicly accessible document-to-text conversion tool\footnote{\href{https://console.upstage.ai/docs/capabilities/document-digitization/document-parsing}{Upstage's Document Parse v240910} was used in this work.}. 

\subsection{Data Preprocessing}
As CC-MAIN offers a broad coverage in both language and document type, filtering is necessary to isolate high-quality English content from CC-MAIN.
To achieve this, we employ FastText's language identification model (lid.176.bin) to detect the primary language of each CC-MAIN document based on its textual content.
Only those confidently classified as English are retained for subsequent stages of the pipeline, ensuring that the English subset of CReSt remains linguistically consistent.

Once the language-consistent documents are identified, each is parsed into the HTML and plain-text formats, the two most commonly used representation of document text. First, for HTML format, the documents are processed using a document-to-text converter. This tool extracts not only the raw textual content, but also preserves essential layout and structural metadata, such as font size, text positioning, and HTML element categories inferred from the visual document.
Next, for plain-text formats, each of the HTML representation is converted into its plain-text equivalent using the \textit{html2text} package\footnote{\url{https://github.com/Alir3z4/html2text}}. This results in dual-format representations: HTML and plain text, for every document in the dataset. By maintaining both formats, CReSt better reflects the diversity and complexity of document-augmented QA scenarios encountered in practical RAG applications.

Following textualization, both the HTML and plain-text representations are segmented into randomly sized chunks, ranging from 2,048 to 16,384 in length. This chunking strategy is designed to simulate the natural variability of document lengths encountered in real-world applications, while also supporting chunk-based retrieval and generation tasks. The resulting preprocessed chunks are stored in a database and serve as the foundation for downstream stages of the dataset construction, including question generation, and negative document sampling.

\subsection{QA Generation}
Based on the HTML and plain-text chunks extracted from the source documents, CReSt adopts a multi-stage QA generation curriculum designed to systematically produce both `simple' and `complex' reasoning question-answer (QA) pairs. This curriculum facilitates a comprehensive evaluation of a LLM’s document-based RAG capabilities by covering a broad range of reasoning scenarios. Before describing the stages of the QA generation process, we first formalize the definition of a reasoning QA pair in CReSt and introduce the criteria used to distinguish between `simple' and `complex' reasoning instances.

In CReSt, a reasoning QA pair refers to a question that requires one or more of the following reasoning skills to arrive at a correct answer: numerical reasoning, tabular reasoning, multi-constraint reasoning, temporal reasoning, and format reasoning—adapted from the reasoning taxonomy introduced in \citet{krishna-2025-frames}. In addition, we introduce textual reasoning, defined as reasoning that requires advanced reading comprehension of the semantics conveyed in the given document. A QA pair is categorized as `simple' if it involves only a single type of reasoning from the list above, whereas it is considered `complex' if it requires combination of different reasoning types.

Generating reasoning-oriented QA pairs entirely by hand is both costly and difficult to scale, while relying solely on model-based generation introduces a risk of factual inaccuracies and reasoning errors. To address these challenges, CReSt adopts a multi-stage QA generation curriculum that guides the generation process in a structured and systematic manner. This curriculum not only facilitates the production of diverse simple and complex QA pairs, but also enables greater control over reasoning types and ensures broad coverage of reasoning patterns. To ensure the correctness and reliability of the generated QA pairs, each example is subsequently validated through human verification.

The curriculum consists of four stages: Key-value pair generation, BasicQA generation, SimpleQA generation, and ComplexQA generation. The core principle of this pipeline is to begin by extracting key information or evidence from the source documents, then formulating basic (query-and-fetch) natural language questions. These basic questions serve as the foundation for the SimpleQA stage, where single reasoning type is applied. Finally, multiple simple reasoning question-answer pairs are combined and composed to construct ComplexQA pairs that require multi-type reasoning to arrive at the correct answer.

QA pairs were generated using GPT-4o~\cite{openai2024gpt4o}, following a multi-stage QA generation curriculum implemented in a multi-turn conversational format. The model is prompted iteratively across different stages of the pipeline to progressively generate question-answer pairs of increasing complexity. The full set of prompts used in each stage is provided in the \cref{subsec:prompts-for-qa-gen}.

At the beginning of the curriculum, for each generation instance, 1 to 5 chunks are randomly sampled from each document, forming the contextual basis for key-value pair generation, which serves as the foundation for downstream QA construction. Each multi-turn conversation is conducted separately for both the HTML and plain-text representations of the selected document chunks.

\subsection{Negative Document Chunk Retrieval}
After generating simple and complex reasoning QA pairs, each conditioned on 1 to 5 chunks from a single document, negative document chunks are retrieved and appended to simulate real-world retrieval challenges, where similar yet irrelevant content may be included. These negative chunks act as retrieval noise, challenging the model’s ability to ground its answers in the correct evidence.

Negative candidates are retrieved based on semantic similarity in the document embedding space. Specifically, we compute top-\textit{k} nearest neighbors using embeddings generated by a publicly available embedding model\footnote{\href{https://console.upstage.ai/docs/capabilities/embeddings}{Upstage's  Embedding model} was used in this work.}. For each QA instance, the value of \textit{k} is set such that the combined number of positive (relevant) and negative (irrelevant) chunks sums to 10, ensuring a consistent retrieval setting across all examples.

\subsection{Human Verification}
To ensure the quality, factual correctness, and evidence alignment of the generated QA pairs, each instance in CReSt undergoes a final human verification step. Human annotators are presented with the question, its corresponding answer, and the set of document chunks (both positive and negative) used during generation.
For each QA pair, two types of verifications are performed concurrently: (1) \textit{Answerability check}, which determines whether the question can be answered based on the provided document chunks; and (2) \textit{Answer and evidence validation}, where annotators assess the correctness of the QA pair and provide citation labels for the corresponding evidence in the document chunks.

\textbf{Answerability Check} Annotators assess whether the question can be accurately answered using the provided document chunks. If insufficient information is available to support a valid answer, the QA pair is labeled as unanswerable. These unanswerable QA pairs are subsequently used to evaluate the model’s refusal capability. To improve data curation efficiency, annotators are instructed to make minor corrections to the question, when appropriate, if such adjustments render the previously unanswerable QA pair answerable.


\textbf{Answer and Evidence Validation} Annotators verify the alignment between the question and answer by reviewing all document chunks (without access to their positive or negative labels). If the question or answer is found to be inaccurate, misaligned or require correction, annotators make the appropriate revision to the QA pair. Additionally, during the verification or revision process, annotators assign evidence labels to identify the supporting chunk(s) from the set of document chunks. This step not only verifies alignment between the answer and its supporting chunk but also facilitates evaluation of the model’s citation capability.
  






\section{Benchmarks and Evaluations}

CReSt provides a holistic evaluation of key capabilities
required in practical RAG scenarios, including answer correctness, accurate citation, and appropriate refusal when information is insufficient.

\subsection{Metrics}

In our dataset, answers are free-form texts rather than a single word or a selection from multiple choices. Therefore, automated lexical overlap metrics (e.g., Lexicon F1 or BLEU) are not suitable for evaluating this task. Instead, to comprehensively assess the model’s answer quality, we adopt an \textit{LLM-as-a-judge} \citep{zheng2023} framework. Specifically, given a gold answer and a model-predicted answer, we query an external LLM to evaluate the semantic equivalence between them. The answers are classified as \textbf{Correct} (fully aligned with the gold answer), \textbf{Partially Correct} (contains only part of the required information), or \textbf{Wrong} (missing essential content or contradictory). This 3-way judgment scheme is illustrated in \cref{subsec:prompt-for-llm-eval}.

To assess the model’s ability to handle unanswerable questions, we explicitly prompt it to generate a pre-defined refusal statement—\textit{I cannot answer because the question is unanswerable with the documents.} when the information provided is not sufficient to answer the question, following the approach introduced by \citep{chen2024rgb}. We then use the presence or absence of this statement to evaluate \textbf{Refusal Accuracy}.

From the perspective of evaluating refusal ability, it is equally important in practical scenarios that models do not refuse when sufficient context is provided. Therefore, we propose an unified evaluation metric that assesses the model’s capability across both answerable and unanswerable cases. We define six possible scenarios based on the gold and predicted response types, and assign a \textbf{Unified Score} to each case as shown in \cref{tab:unified_score}. This scoring scheme ensures that overly conservative models, those that frequently refuse to answer, do not receive high scores, as refusals in answerable cases are penalized. We compute it as the arithmetic mean of the non-refusal and refusal scores.


\begin{wraptable}{r}{0.4\textwidth}
  \centering
  \small
  \caption{Unified Score scoring scheme based on gold and predicted outputs}
  \label{tab:unified_score}
  \begin{tabular}{llr}
    \toprule
    Gold          & Predicted           & Score \\ 
    \midrule
    Non-Refusal   & Correct             & 1.0   \\
    Non-Refusal   & Partially Correct   & 0.5   \\
    Non-Refusal   & Wrong               & 0.0   \\
    Non-Refusal   & Refusal             & -1.0  \\
    \midrule
    Refusal       & Refusal             & 1.0   \\
    Refusal       & Non-Refusal         & 0.0   \\
    \bottomrule
  \end{tabular}
\end{wraptable}

In addition to answer correctness and refusal handling, the ability to identify the document containing the supporting evidence is also crucial in practical RAG scenarios. We define citation ability as the model’s capacity to generate source references (e.g., [1], [2], etc.) appended to the answer, indicating the corresponding document chunk(s) containing the evidence. We evaluate the citation ability using \textit{Citation Precision} $= \frac{|P \cap G|}{|P|}$ and \textit{Citation Recall} $= \frac{|P \cap G|}{|G|}$ as introduced by \citet{gao2023}. Unlike their NLI-based method, we leverage gold citation annotations and compute these metrics through direct set-based comparison. Specifically, let $G$ denote the set of gold citations and $P$ denote the set of predicted citations.
\subsection{Experimental Setups}

In our experiments, we employed a diverse set of open-source and proprietary models. For the open-source models, we utilized the Qwen2.5 \citep{qwen2024} series, which includes models of various sizes: 3B, 7B, 14B, 32B, and 72B. We also included Llama-3.3 70B \citep{llama2024} as a representative open-source model. Among proprietary models, we selected GPT-4o (gpt-4o-2024-08-06)\citep{openai2024gpt4o} and GPT-4.1 (gpt-4.1-2025-04-14) \citep{openai2025gpt41}, as well as o3-mini (o3-mini-2025-01-31)\citep{openai2025o3mini} and o4-mini (o4-mini-2025-04-16) \citep{openai2025o4mini}, which have recently attracted attention for their strong reasoning capabilities.

We used the GPT-4o (gpt-4o-2024-08-06) as the judge model for the LLM-as-a-judge. The prompts used for both model inference and evaluation are provided in the \cref{subsec:prompts-for-inference}. As shown in the prompts, all models were instructed to follow a Chain-of-Thought (CoT) \citep{wei2022} reasoning format. However, during evaluation, we considered only the final answer enclosed between \texttt{<Answer>} and \texttt{</Answer>} tags, excluding the intermediate reasoning process from scoring.

\subsection{Evaluation Results}

\subsubsection{Answer Correctness}

\begin{table}[ht]
\centering
\caption{Answer correctness results across English and Korean. Here, C, P, and W denote the rates of Correct, Partially Correct, and Wrong answers, respectively.}
\label{tab:answer_correctness}
\resizebox{\textwidth}{!}{%
\scriptsize
\begin{tabular}{l|c|c|c|c|c|c}
\toprule
\multirow{2}{*}{} & \multicolumn{3}{c|}{\textbf{English}} & \multicolumn{3}{c}{\textbf{Korean}} \\
\cmidrule(lr){2-4} \cmidrule(lr){5-7}
 & Unified Score & Non-Refusal (C/P/W) & Refusal Accuracy & Unified Score & Non-Refusal (C/P/W) & Refusal Accuracy \\
\midrule
\textbf{Qwen2.5-3B-Instruct} & 0.1756 & 10.84\% / 44.68\% / 44.49\% & 6.13\% & 0.1645 & 3.40\% / 53.68\% / 42.92\% & 9.16\% \\
\textbf{Qwen2.5-7B-Instruct} & 0.2482 & 22.81\% / 36.88\% / 40.30\% & 28.16\% & 0.2037 & 10.20\% / 52.12\% / 37.68\% & 9.57\% \\
\textbf{Qwen2.5-14B-Instruct} & 0.2730 & 29.64\% / 40.73\% / 29.64\% & 8.24\% & 0.2513 & 19.03\% / 53.12\% / 27.84\% & 6.52\% \\
\textbf{Qwen2.5-32B-Instruct} & 0.3650 & 27.19\% / 43.92\% / 28.90\% & \textbf{36.21\%} & 0.3309 & 17.56\% / 57.22\% / 25.21\% & 30.35\% \\
\textbf{Qwen2.5-72B-Instruct} & 0.3074 & 37.83\% / 39.92\% / 22.24\% & 8.62\% & 0.2999 & 24.08\% / 61.05\% / 14.87\% & 6.52\% \\
\textbf{Llama-3.3-70B-Instruct} & 0.3253 & 23.95\% / 47.91\% / 28.14\% & 31.23\% & 0.2598 & 12.04\% / 69.97\% / 17.99\% & 6.92\% \\
\textbf{GPT-4o} & 0.3777 & 34.79\% / 40.68\% / 24.52\% & 32.95\% & 0.3841 & 24.93\% / 56.94\% / 18.13\% & \textbf{33.20\%} \\
\textbf{GPT-4.1} & 0.3679 & 46.77\% / 35.55\% / 17.68\% & 12.64\% & 0.3826 & 43.63\% / 46.74\% / 9.63\% & 10.79\% \\
\textbf{o3-mini} & 0.3870 & 59.70\% / 28.14\% / \textbf{12.17\%} & 4.21\% & 0.3753 & \textbf{50.28\%} / 42.63\% / \textbf{7.08\%} & 3.46\% \\
\textbf{o4-mini} & \textbf{0.4390} & \textbf{59.89\%} / 24.71\% / 15.40\% & 20.88\% & \textbf{0.4458} & 47.88\% / 42.78\% / 9.35\% & 23.01\% \\
\bottomrule
\end{tabular}
}
\vspace{1pt}
\end{table}

The results, summarized in \cref{tab:answer_correctness}, present the correctness of the answer on different models.
Overall, proprietary models, such as GPT and o-series, consistently outperform open-source models in both English and Korean settings.
However, even the best-performing model, o4-mini, achieves a unified score of only 0.43 to 0.44, indicating that the benchmark presents genuinely challenging cases that remain difficult even for state-of-the-art (SoTA) models.
When compared with a non-reasoning model like GPT-4.1, its superior performance highlights that better performance is achieved on the CReSt benchmark when equipped with strong reasoning capabilities.
This reinforces the benchmark's alignment to practical document RAG applications, where advanced reasoning is often essential.

Beyond the Unified Scores, the breakdown into Non-Refusal and Refusal accuracy offers deeper insights when comparing model performance. For example, although GPT-4o and GPT-4.1 achieve similar Unified Scores, a notable gap is observed in their Refusal Accuracy. GPT-4o demonstrates strong ability in identifying unanswerable questions, yet its accuracy on answerable cases is comparatively lower. Conversely, GPT-4.1 performs better on answerable questions but struggles more with refusal cases.
The importance of the balance is further reflected with Qwen model series.
Although there is a general trend of improved performance with increasing model size, the 32B model outperforms the 72B model in terms of the Unified Score, despite exhibiting a higher wrong rate in the non-refusal setting.
A closer examination suggests that although the 72B model achieved higher scores in non-refusal cases, its limited ability to accurately capture refusal cases results in a lower overall score.
This highlights a key aspect of our benchmark, where performance in both refusal and non-refusal scenarios is essential and should be evaluated jointly.

In the aspect of language robustness, LLaMA 3.3–70B shows a considerable performance gap between English and Korean, which is a comparatively low-resource language.
This underscores the need for language-specific evaluation and positions our benchmark as a key tool for developing Korean-capable models.

\subsubsection{Citation Performance}
\begin{table}[ht]
\centering
\scriptsize
\caption{Citation performance across models.}
\label{tab:citation_precision_recall}
\begin{tabular}{l|ccc|ccc}
\toprule
\multirow{2}{*}{} & \multicolumn{3}{c|}{\textbf{English}} & \multicolumn{3}{c}{\textbf{Korean}} \\
\cmidrule(lr){2-4} \cmidrule(lr){5-7}
 & Precision & Recall & F1 & Precision & Recall & F1\\
\midrule
\textbf{Qwen2.5-3B-Instruct} & 5.24\% & 7.86\% & 6.29\% & 6.60\% & 11.00\% & 8.25\% \\
\textbf{Qwen2.5-7B-Instruct} & 42.06\% & 55.47\% & 47.84\% & 35.24\% & 54.41\% & 42.78\% \\
\textbf{Qwen2.5-14B-Instruct} & 32.65\% & 36.61\% & 34.52\% & 46.85\% & 56.23\% & 51.11\% \\
\textbf{Qwen2.5-32B-Instruct} & 63.47\% & 74.09\% & 68.37\% & 61.39\% & 71.29\% & 65.97\% \\
\textbf{Qwen2.5-72B-Instruct} & 67.63\% & 78.32\% & 72.58\% &61.71\% & 75.71\% & 68.00\% \\
\textbf{Llama-3.3-70B-Instruct} & 30.85\% & 46.19\% & 36.99\% & 32.54\% & 54.95\% & 40.87\% \\
\textbf{GPT-4o} & 45.53\% & 57.33\% & 50.75\% & 53.18\% & 74.94\% & 62.21\% \\
\textbf{GPT-4.1} & 67.81\% & \textbf{84.86\%} & 75.38\% & 66.17\% & \textbf{92.73\%} & 77.23\% \\
\textbf{o3-mini} & 76.19\% & 84.27\% & \textbf{80.03\%} & 74.26\% & 82.50\% & 78.16\% \\
\textbf{o4-mini} & \textbf{77.82\%} & 80.93\% & 79.34\% & \textbf{75.53\%} & 83.01\% & \textbf{79.09\%} \\
\bottomrule
\end{tabular}
\vspace{1pt}
\end{table}

Citation is particularly important in RAG scenarios, where factual grounding is crucial.
As shown in \cref{tab:citation_precision_recall}, citation capability generally correlates with answer correctness: where models that cite relevant evidence better tend to achieve higher answer accuracy.

However, an interesting discrepancy emerges when comparing Qwen2.5-32B-Instruct and Qwen2.5-72B-Instruct. Although the 72B model achieves higher citation scores, its overall Unified Score is lower than that of the 32B model. This suggests that while the 72B model is more effective at identifying relevant documents, it struggles to accurately distinguish between answerable and unanswerable queries, ultimately lowering its overall performance.
This case illustrates that, although citation capability is essential in RAG scenarios, it must be assessed alongside other critical dimensions to enable a truly holistic and meaningful comparison of model performance.



\subsubsection{Performance across Different Levels of Difficulty}
\begin{table}[ht]
\centering
\caption{Performances across reasoning difficulty levels (C: Correct, P: Partially Correct, W: Wrong).}
\label{tab:difficulty_comparison}
\resizebox{\textwidth}{!}{%
\begin{tabular}{l|c|c|c|c|c|c}
\toprule
\multirow{2}{*}{} & \multicolumn{3}{c|}{\textbf{SimpleQA}} & \multicolumn{3}{c}{\textbf{ComplexQA}} \\
\cmidrule(lr){2-4} \cmidrule(lr){5-7}
 & Unified Score & Non-Refusal (C/P/W) & Refusal & Unified Score & Non-Refusal (C/P/W) & Refusal \\
\midrule
\textbf{Qwen2.5-72B-Instruct}     & 0.3697 & 54.71\% / 29.60\% / 15.70\% & 9.38\%  & 0.2630 & 25.41\% / 47.52\% / 27.06\% & 8.38\% \\
\textbf{Llama-3.3-70B-Instruct}   & 0.3751 & 40.36\% / 39.91\% / 19.73\% & 21.88\% & 0.2695 & 11.88\% / 53.80\% / 34.32\% & \textbf{34.26\%} \\
\textbf{GPT-4o}          & 0.4366 & 51.12\% / 28.25\% / 20.63\% & \textbf{29.69\%} & 0.3276 & 22.77\% / 49.83\% / 27.39\% & 34.01\% \\
\textbf{GPT-4.1}         & 0.4072 & 56.95\% / 27.35\% / 15.70\% & 14.84\% & 0.3434 & 39.27\% / 41.58\% / 19.14\% & 11.93\% \\
\textbf{o4-mini}         & \textbf{0.4493} & \textbf{71.75\%} / 16.59\% / 11.66\% & 12.50\% & \textbf{0.4142} & \textbf{51.16\%} / 30.69\% / 18.15\% & 23.60\% \\
\bottomrule
\end{tabular}
}
\end{table}
\cref{tab:difficulty_comparison} shows the performance across the difficulty levels of the task. As expected, performance generally declines on ComplexQA compared to SimpleQA, highlighting the increased challenge posed by deeper reasoning requirements. Notably, the performance gap between o4-mini and GPT-4o widens under complex tasks, suggesting that models with stronger reasoning capabilities, such as o4-mini, exhibit greater robustness when faced with more demanding reasoning scenarios.

\subsubsection{Performance across Reasoning Types}

\begin{figure}[h]
\centering
\includegraphics[width=0.7\textwidth]{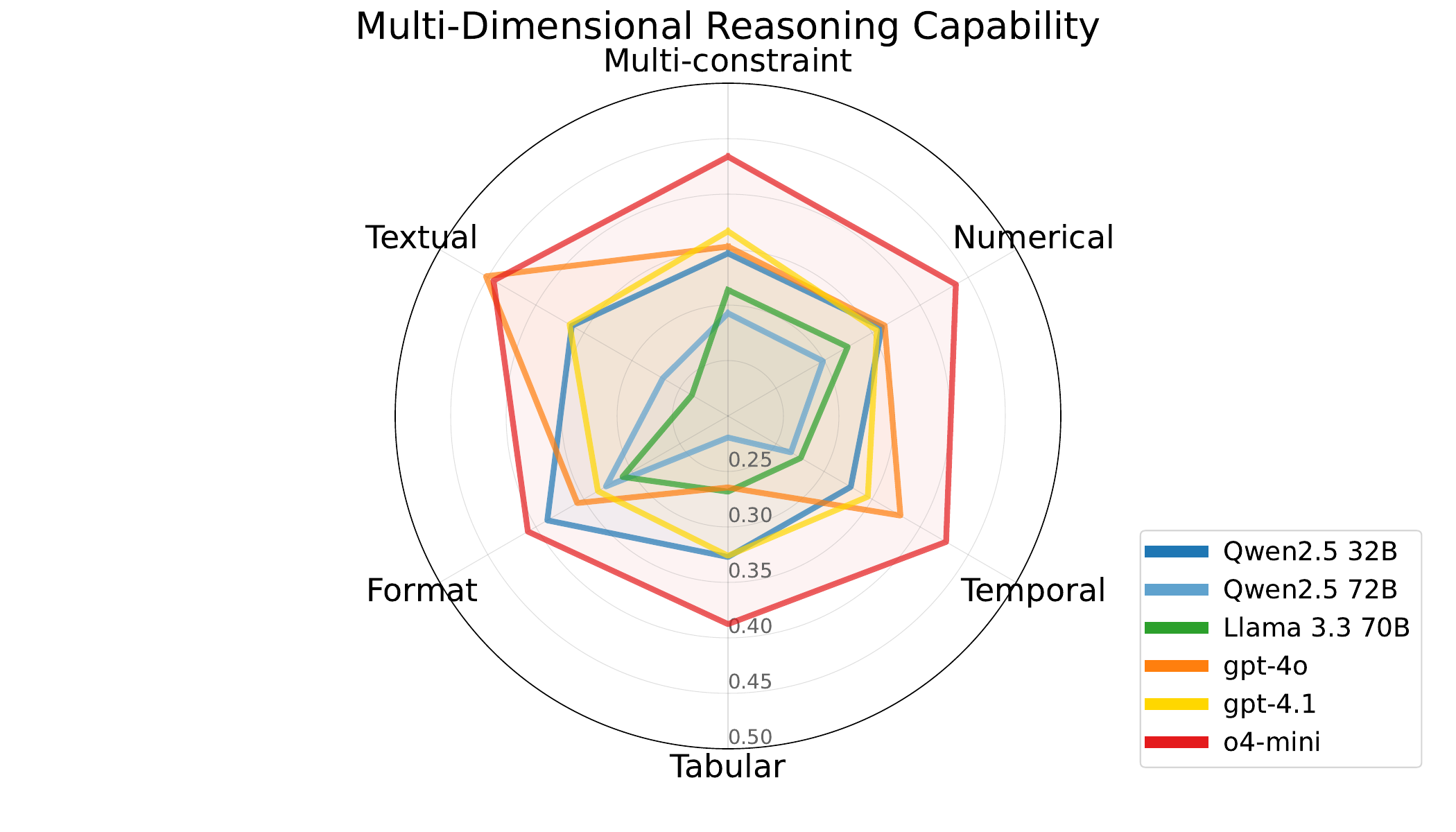}
\caption{Reasoning capability comparison across types.}
\label{fig:reasoning_radar}
\end{figure}

Reasoning types provided in CReSt enable focused analysis of model performance across distinct reasoning categories.
As shown in \cref{fig:reasoning_radar}, o4-mini demonstrates strong and balanced performance across all reasoning categories. In contrast, other models show greater variability depending on the reasoning type. For example, both GPT-4o and Qwen2.5-72B-Instruct underperform in tabular reasoning, while o4-mini excels in numerical reasoning and GPT-4o shows particular strength in textual reasoning.

\subsubsection{Inference Methods}

\begin{table}[h]
\centering
\scriptsize
\caption{Performance comparison across different inference methods for gpt-4o-mini and gpt-4o models in English dataset.
Direct Answer excels in conservative refusal accuracy, Least-to-Most achieves the highest overall performance, and other methods boost partial correctness but incur more errors, highlighting a trade-off between reasoning depth and accuracy.
}
\label{tab:test-time-methods}
\begin{tabular}{l|c|c|c}
\toprule
 & Unified & Non-Refusal (C/P/W) & Refusal Accuracy \\
\midrule
\textbf{CoT (Baseline)} & & & \\
gpt-4o-mini & 0.3276 & 26.47\% / 47.45\% / 26.09\% & 26.10\% \\
gpt-4o & 0.3777 & 34.79\% / 40.68\% / 24.52\% & 32.95\% \\
Llama-3.3-70B-Instruct & 0.3844 & 26.98\% / 47.17\% / 25.85\% & 31.23\% \\
Qwen2.5-32B-Instruct & 0.3478 & 32.45\% / 40.75\% / 26.79\% & 26.15\% \\
\midrule

\textbf{Direct Answer} & & & \\
gpt-4o-mini & 0.3123 & 21.51\% / 16.23\% / 62.26\% & \textbf{87.93\%} \\
gpt-4o & 0.4030 & 35.17\% / 26.43\% / 38.40\% & 60.92\% \\
Llama-3.3-70B-Instruct & 0.2778 & 16.98\% / 26.98\% / 56.04\% & 74.90\% \\
Qwen2.5-32B-Instruct & 0.3037 & 17.55\% / 28.68\% / 53.77\% & 74.33\% \\
\midrule

\textbf{CoD} & & & \\
gpt-4o-mini & 0.3288 & 23.96\% / 27.74\% / 48.30\% & 63.22\% \\
gpt-4o & 0.3677 & 30.94\% / 39.06\% / 30.00\% & 41.38\% \\
Llama-3.3-70B-Instruct & 0.3272 & \textbf{39.56\%} / 33.33\% / 27.11\% & 48.47\% \\
Qwen2.5-32B-Instruct & 0.3647 & 26.60\% / 43.21\% / 30.19\% & 38.51\% \\
\midrule

\textbf{Plan-And-Solve} & & & \\
gpt-4o-mini & 0.3299 & 27.36\% / 46.60\% / 26.04\% & 24.38\% \\
gpt-4o & 0.3717 & 36.42\% / 45.85\% / \textbf{17.74\%} & 19.35\% \\
Llama-3.3-70B-Instruct & 0.3198 & 22.83\% / 48.49\% / 28.68\% &  31.61\% \\
Qwen2.5-32B-Instruct & 0.3436 & 33.40\% / 42.83\% / 23.77\% &  21.26\% \\
\midrule

\textbf{Least-to-Most} & & & \\
gpt-4o-mini & 0.3666 & 29.06\% / 29.81\% / 41.13\% & 57.85\% \\
gpt-4o & \textbf{0.4502} & 38.49\% / 29.62\% / 31.89\% & 59.00\% \\
Llama-3.3-70B-Instruct & 0.4001 & 33.40\% / 37.17\% / 29.43\% & 45.02\% \\
Qwen2.5-32B-Instruct & 0.3638 & 33.77\% / 28.87\% / 37.36\% & 51.15\% \\
\bottomrule
\end{tabular}
\end{table}

Alongside model comparisons, we further evaluate recent advancements in inference methods to assess their effectiveness in our benchmark.
The baseline used in prior experiments is the Chain-of-Thought (CoT) method \citep{wei2022}.
For comparison, we include a Direct Answer setting, where models are instructed to respond immediately without engaging in any explicit reasoning process.

We also explore several alternative approaches. Chain-of-draft (CoD) \citep{xu2025cod} improves efficiency over CoT by replacing the thought process with a brief draft consisting of only a few words. Plan-And-Solve method \citep{wang-etal-2023-plan-and-solve} similarly prompts step-by-step reasoning but first instructs the model to formulate a plan before executing it to solve the problem. Least-to-Most (L2M) method \citep{zhou2023leasttomost} takes a decomposition-based approach, guiding the model to break the original task into smaller sub-problems and solve them sequentially.

\cref{tab:test-time-methods} presents distinct performance patterns across inference methods when applied to various LLMs. Among them, L2M approach combined with GPT-4o demonstrates the highest overall performance, underscoring its effectiveness in decomposing complex queries into manageable subproblems. This observation might stem from the alignment with dataset construction process, where incremental question generation naturally corresponds to the decomposition strategy in inference.

Interestingly, simpler inference methods, such as Direct Answer, tend to achieve higher refusal accuracy, reflecting a conservative response strategy that prioritizes avoiding incorrect outputs.
However, this comes at the expense of overall response quality, as seen in the lower unified scores. The fact that L2M also achieves relatively strong refusal accuracy while maintaining high response quality suggests that our benchmark rewards more advanced reasoning strategies that effectively balance both refusal handling and accurate response generation.
Additionally, methods like CoT and Plan-and-Solve still exhibit notable reductions in wrong responses within non-refusal situation, indicating that structured reasoning steps can mitigate reasoning errors.

Model-specific trends reveal that GPT-4o performs unusually well under Direct Answer, maintaining a rare balance between refusal and response accuracy. This trend suggests that GPT-4o's capacity to handle both aspects effectively under a straightforward usage. Meanwhile, open-source models struggle under Direct Answer, with a sharp drop in overall scores. These models seem to rely on structured reasoning to mitigate errors, as seen in their relatively stable performance under CoT and Plan-And-Solve.
This analysis underscores the importance of selecting inference methods that align with both the model’s reasoning capabilities.

\section{Conclusion}
In this paper, we propose CReSt, a benchmark designed to holistically evaluate the diverse capabilities required of LLMs for developing real-world RAG (Retrieval-Augmented Generation) applications. We construct a dataset in both English and Korean that requires complex reasoning, grounded in a diverse set of documents representative of real-world use cases. Based on this dataset, we design the CReSt benchmark to evaluate model capabilities in realistic RAG scenarios. Through extensive experiments, we demonstrate that CReSt effectively reveals the strengths and weaknesses of various models in practical settings, offering valuable insights. We believe our benchmark will serve as a useful resource for future research and production-level development of RAG systems.
\section{Limitations}
Although CReSt contributes significant advances in the development of real-world RAG applications, several limitations are remain. First, our benchmark is designed to evaluate the capabilities of LLMs within RAG applications, rather than to assess the end-to-end performance of fully integrated RAG pipelines. As such, evaluating the overall performance of the complete RAG pipeline, including retrieval, ranking, and generation modules, is beyond the scope of this study.
Second, our experiments reveal that model performance is highly sensitive to prompt design. While this highlights the potential for further improvements through prompt optimization, we do not explore rigorous prompt engineering strategies in this work and leave such investigations for future research.
Third, although our dataset includes phrase level annotations for citations, for the sake of simplicity, we evaluate citation performance at the document level in this study. In future work, we plan to leverage these fine-grained annotations to conduct more detailed and precise evaluations of citation accuracy.
Finally, although we aim to support real-world application development by utilizing diverse documents and QA pairs, the dataset may still lack sufficient coverage of the full range of real-world scenarios and language diversity. Expanding the dataset to address these limitations is also left for future work.
\bibliographystyle{plainnat}
\bibliography{refs}

\begin{thebibliography}{25}
\providecommand{\natexlab}[1]{#1}
\providecommand{\url}[1]{\texttt{#1}}
\expandafter\ifx\csname urlstyle\endcsname\relax
  \providecommand{\doi}[1]{doi: #1}\else
  \providecommand{\doi}{doi: \begingroup \urlstyle{rm}\Url}\fi

\bibitem[Alibaba(2024)]{qwen2024}
Alibaba.
\newblock Qwen2.5: A party of foundation models!
\newblock \url{https://qwenlm.github.io/blog/qwen2.5/}, 2024.

\bibitem[Chen et~al.(2024)Chen, Lin, Han, and Sun]{chen2024rgb}
Jiawei Chen, Hongyu Lin, Xianpei Han, and Le~Sun.
\newblock Benchmarking large language models in retrieval-augmented generation.
\newblock In \emph{Proceedings of the Thirty-Eighth AAAI Conference on Artificial Intelligence and Thirty-Sixth Conference on Innovative Applications of Artificial Intelligence and Fourteenth Symposium on Educational Advances in Artificial Intelligence}, pages 17754--17762, 2024.
\newblock \doi{10.1609/aaai.v38i16.29728}.
\newblock URL \url{https://doi.org/10.1609/aaai.v38i16.29728}.

\bibitem[Fan et~al.(2024)Fan, Ding, Ning, Wang, Li, Yin, Chua, and Li]{fan2024survey}
Wenqi Fan, Yujuan Ding, Liangbo Ning, Shijie Wang, Hengyun Li, Dawei Yin, Tat{-}Seng Chua, and Qing Li.
\newblock A survey on rag meeting llms: Towards retrieval-augmented large language models.
\newblock In \emph{Proceedings of the 30th ACM SIGKDD Conference on Knowledge Discovery and Data Mining (KDD '24)}, pages 6491--6501. ACM, 2024.
\newblock \doi{10.1145/3637528.3671470}.

\bibitem[Gao et~al.(2023{\natexlab{a}})Gao, Yen, Yu, and Chen]{gao-2023-alce}
Tianyu Gao, Howard Yen, Jiatong Yu, and Danqi Chen.
\newblock Enabling large language models to generate text with citations.
\newblock In \emph{Proceedings of the 2023 Conference on Empirical Methods in Natural Language Processing}, pages 6465--6488, Singapore, December 2023{\natexlab{a}}. Association for Computational Linguistics.
\newblock \doi{10.18653/v1/2023.emnlp-main.398}.
\newblock URL \url{https://aclanthology.org/2023.emnlp-main.398/}.

\bibitem[Gao et~al.(2023{\natexlab{b}})Gao, Yen, Yu, and Chen]{gao2023}
Tianyu Gao, Howard Yen, Jiatong Yu, and Danqi Chen.
\newblock Enabling large language models to generate text with citations.
\newblock In \emph{Proceedings of the 2023 Conference on Empirical Methods in Natural Language Processing (EMNLP)}, Singapore, 2023{\natexlab{b}}. Association for Computational Linguistics.

\bibitem[Gao et~al.(2023{\natexlab{c}})Gao, Xiong, Gao, Jia, Pan, Bi, Dai, Sun, Wang, and Wang]{gao2023rag}
Yunfan Gao, Yun Xiong, Xinyu Gao, Kangxiang Jia, Jinliu Pan, Yuxi Bi, Yi~Dai, Jiawei Sun, Meng Wang, and Haofen Wang.
\newblock Retrieval-augmented generation for large language models: A survey.
\newblock \emph{arXiv preprint arXiv:2312.10997}, 2023{\natexlab{c}}.
\newblock URL \url{https://arxiv.org/abs/2312.10997}.

\bibitem[Hui et~al.(2024)Hui, Lu, and Zhang]{hui2024uda}
Yulong Hui, Yao Lu, and Huanchen Zhang.
\newblock Uda: A benchmark suite for retrieval augmented generation in real-world document analysis.
\newblock In A.~Globerson, L.~Mackey, D.~Belgrave, A.~Fan, U.~Paquet, J.~Tomczak, and C.~Zhang, editors, \emph{Advances in Neural Information Processing Systems}, volume~37, pages 67200--67217. Curran Associates, Inc., 2024.
\newblock URL \url{https://proceedings.neurips.cc/paper_files/paper/2024/file/7c06759d1a8567f087b02e8589454917-Paper-Datasets_and_Benchmarks_Track.pdf}.

\bibitem[Krishna et~al.(2025)Krishna, Krishna, Mohananey, Schwarcz, Stambler, Upadhyay, and Faruqui]{krishna-2025-frames}
Satyapriya Krishna, Kalpesh Krishna, Anhad Mohananey, Steven Schwarcz, Adam Stambler, Shyam Upadhyay, and Manaal Faruqui.
\newblock Fact, fetch, and reason: A unified evaluation of retrieval-augmented generation.
\newblock In \emph{Proceedings of the 2025 Conference of the North American Chapter of the Association for Computational Linguistics: Human Language Technologies (Volume 1: Long Papers)}, pages 4745--4759, Albuquerque, New Mexico, April 2025. Association for Computational Linguistics.
\newblock URL \url{https://aclanthology.org/2025.naacl-long.243/}.

\bibitem[Lyu et~al.(2024)Lyu, Li, Niu, Xiong, Tang, Wang, Wu, Liu, Xu, and Chen]{lyu2024crudrag}
Yuanjie Lyu, Zhiyu Li, Simin Niu, Feiyu Xiong, Bo~Tang, Wenjin Wang, Hao Wu, Huanyong Liu, Tong Xu, and Enhong Chen.
\newblock {CRUD-RAG}: A comprehensive chinese benchmark for retrieval-augmented generation of large language models.
\newblock \emph{ACM Transactions on Information Systems}, 43\penalty0 (2):\penalty0 1--32, 2024.
\newblock \doi{10.1145/3701228}.
\newblock URL \url{https://doi.org/10.1145/3701228}.

\bibitem[Meta(2024)]{llama2024}
Meta.
\newblock Llama 3.3.
\newblock \url{https://www.llama.com/docs/model-cards-and-prompt-formats/llama3_3/}, 2024.

\bibitem[OpenAI(2024)]{openai2024gpt4o}
OpenAI.
\newblock Hello gpt-4o.
\newblock \url{https://openai.com/index/hello-gpt-4o/}, 2024.

\bibitem[OpenAI(2025{\natexlab{a}})]{openai2025gpt41}
OpenAI.
\newblock Introducing gpt-4.1 in the api.
\newblock \url{https://openai.com/index/gpt-4-1/}, 2025{\natexlab{a}}.

\bibitem[OpenAI(2025{\natexlab{b}})]{openai2025o3mini}
OpenAI.
\newblock Openai o3-mini.
\newblock \url{https://openai.com/index/openai-o3-mini/}, 2025{\natexlab{b}}.

\bibitem[OpenAI(2025{\natexlab{c}})]{openai2025o4mini}
OpenAI.
\newblock Introducing openai o3 and o4-mini.
\newblock \url{https://openai.com/index/introducing-o3-and-o4-mini/}, 2025{\natexlab{c}}.

\bibitem[OpenAI et~al.(2024)OpenAI, Achiam, Adler, Agarwal, Ahmad, Akkaya, Aleman, Almeida, Altenschmidt, Altman, Anadkat, Avila, Babuschkin, Balaji, Balcom, Baltescu, Bao, Bavarian, Belgum, et~al.]{gpt4}
OpenAI, Josh Achiam, Steven Adler, Sandhini Agarwal, Lama Ahmad, Ilge Akkaya, Florencia~Leoni Aleman, Diogo Almeida, Janko Altenschmidt, Sam Altman, Shyamal Anadkat, Red Avila, Igor Babuschkin, Suchir Balaji, Valerie Balcom, Paul Baltescu, Haiming Bao, Mohammad Bavarian, Jeff Belgum, et~al.
\newblock Gpt-4 technical report, 2024.
\newblock URL \url{https://arxiv.org/abs/2303.08774}.

\bibitem[Petroni et~al.(2021)Petroni, Piktus, Fan, Lewis, Yazdani, De~Cao, Thorne, Jernite, Karpukhin, Maillard, Plachouras, Rockt{\"a}schel, and Riedel]{petroni-2021-kilt}
Fabio Petroni, Aleksandra Piktus, Angela Fan, Patrick Lewis, Majid Yazdani, Nicola De~Cao, James Thorne, Yacine Jernite, Vladimir Karpukhin, Jean Maillard, Vassilis Plachouras, Tim Rockt{\"a}schel, and Sebastian Riedel.
\newblock {KILT}: a benchmark for knowledge intensive language tasks.
\newblock In \emph{Proceedings of the 2021 Conference of the North American Chapter of the Association for Computational Linguistics: Human Language Technologies}, pages 2523--2544, Online, June 2021. Association for Computational Linguistics.
\newblock \doi{10.18653/v1/2021.naacl-main.200}.
\newblock URL \url{https://aclanthology.org/2021.naacl-main.200/}.

\bibitem[Tan et~al.(2025)Tan, Dou, Wang, Wang, Chen, and Wen]{tan2025htmlrag}
Jiejun Tan, Zhicheng Dou, Wen Wang, Mang Wang, Weipeng Chen, and Ji-Rong Wen.
\newblock Htmlrag: Html is better than plain text for modeling retrieved knowledge in rag systems.
\newblock In \emph{Proceedings of the ACM on Web Conference 2025 (WWW '25)}, pages 1733--1746. ACM, 2025.
\newblock \doi{10.1145/3696410.3714546}.

\bibitem[Tanaka et~al.(2025)Tanaka, Iki, Hasegawa, Nishida, Saito, and Suzuki]{tanaka2025vdocrag}
Ryota Tanaka, Taichi Iki, Taku Hasegawa, Kyosuke Nishida, Kuniko Saito, and Jun Suzuki.
\newblock Vdocrag: Retrieval-augmented generation over visually-rich documents.
\newblock In \emph{CVPR}, 2025.

\bibitem[Thorne et~al.(2018)Thorne, Vlachos, Christodoulopoulos, and Mittal]{thorne-2018-fever}
James Thorne, Andreas Vlachos, Christos Christodoulopoulos, and Arpit Mittal.
\newblock {FEVER}: a large-scale dataset for fact extraction and verification.
\newblock In \emph{Proceedings of the 2018 Conference of the North American Chapter of the Association for Computational Linguistics: Human Language Technologies, Volume 1 (Long Papers)}, pages 809--819, New Orleans, Louisiana, June 2018. Association for Computational Linguistics.
\newblock \doi{10.18653/v1/N18-1074}.
\newblock URL \url{https://aclanthology.org/N18-1074}.

\bibitem[Wang et~al.(2023)Wang, Xu, Lan, Hu, Lan, Lee, and Lim]{wang-etal-2023-plan-and-solve}
Lei Wang, Wanyu Xu, Yihuai Lan, Zhiqiang Hu, Yunshi Lan, Roy Ka-Wei Lee, and Ee-Peng Lim.
\newblock Plan-and-solve prompting: Improving zero-shot chain-of-thought reasoning by large language models.
\newblock In Anna Rogers, Jordan Boyd-Graber, and Naoaki Okazaki, editors, \emph{Proceedings of the 61st Annual Meeting of the Association for Computational Linguistics (Volume 1: Long Papers)}, pages 2609--2634, Toronto, Canada, July 2023. Association for Computational Linguistics.
\newblock \doi{10.18653/v1/2023.acl-long.147}.
\newblock URL \url{https://aclanthology.org/2023.acl-long.147/}.

\bibitem[Wei et~al.(2022)Wei, Wang, Schuurmans, Bosma, Ichter, Xia, Chi, Le, and Zhou]{wei2022}
Jason Wei, Xuezhi Wang, Dale Schuurmans, Maarten Bosma, Brian Ichter, Fei Xia, Ed~Chi, Quoc Le, and Denny Zhou.
\newblock Chain-of-thought prompting elicits reasoning in large language models.
\newblock In \emph{Proceedings of the 36th International Conference on Neural Information Processing Systems}, 2022.

\bibitem[Xu et~al.(2025)Xu, Xie, Zhao, and He]{xu2025cod}
Silei Xu, Wenhao Xie, Lingxiao Zhao, and Pengcheng He.
\newblock Chain of draft: Thinking faster by writing less.
\newblock \emph{arXiv preprint arXiv:2502.18600}, 2025.

\bibitem[Yang et~al.(2018)Yang, Qi, Zhang, Bengio, Cohen, Salakhutdinov, and Manning]{yang-2018-hotpotqa}
Zhilin Yang, Peng Qi, Saizheng Zhang, Yoshua Bengio, William Cohen, Ruslan Salakhutdinov, and Christopher~D. Manning.
\newblock Hotpotqa: A dataset for diverse, explainable multi-hop question answering.
\newblock In \emph{Proceedings of the 2018 Conference on Empirical Methods in Natural Language Processing}, pages 2369--2380, Brussels, Belgium, October-November 2018. Association for Computational Linguistics.
\newblock \doi{10.18653/v1/D18-1259}.
\newblock URL \url{https://aclanthology.org/D18-1259}.

\bibitem[Zheng et~al.(2023)Zheng, Chiang, Sheng, Zhuang, Wu, Zhuang, Lin, Li, Li, Xing, Zhang, Gonzalez, and Stoica]{zheng2023}
Lianmin Zheng, Wei-Lin Chiang, Ying Sheng, Siyuan Zhuang, Zhanghao Wu, Yonghao Zhuang, Zi~Lin, Zhuohan Li, Dacheng Li, Eric~P. Xing, Hao Zhang, Joseph~E. Gonzalez, and Ion Stoica.
\newblock Judging llm-as-a-judge with mt-bench and chatbot arena.
\newblock In \emph{Proceedings of the 37th International Conference on Neural Information Processing Systems}, 2023.

\bibitem[Zhou et~al.(2023)Zhou, Sch{\"a}rli, Hou, Wei, Scales, Wang, Schuurmans, Cui, Bousquet, Le, and Chi]{zhou2023leasttomost}
Denny Zhou, Nathanael Sch{\"a}rli, Le~Hou, Jason Wei, Nathan Scales, Xuezhi Wang, Dale Schuurmans, Claire Cui, Olivier Bousquet, Quoc~V Le, and Ed~H. Chi.
\newblock Least-to-most prompting enables complex reasoning in large language models.
\newblock In \emph{The Eleventh International Conference on Learning Representations}, 2023.
\newblock URL \url{https://openreview.net/forum?id=WZH7099tgfM}.

\end{thebibliography}

\appendix
\newpage
\section{Dataset Statistics}
\label{sec:dataset_statistics}

\begin{figure}[h!]
\centering
\includegraphics[width=0.8\textwidth]{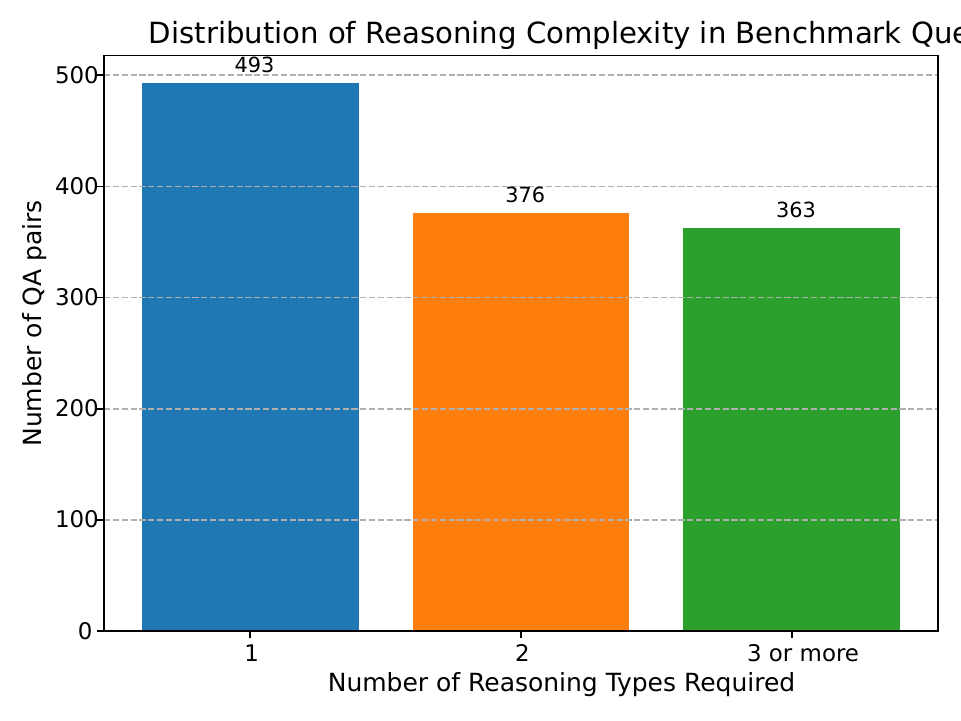}
\caption{Distribution of Reasoning Complexity in Benchmark data. Our benchmark includes various questions with reasoning types ranging from one to three or more, allowing for multidimensional evaluation of model capabilities.}
\label{fig:complexity_dist}
\end{figure}

\cref{tab:dataset_stats} shows the number of examples categorized by refusal status of the answers, language, and difficulty, while \cref{tab:reasoning_types} illustrates the distribution of reasoning types appearing in both SimpleQA and ComplexQA. \cref{fig:complexity_dist} represents the distribution of reasoning types across QA examples.

\begin{table}[h]
\centering
\caption{Number of examples grouped by refusal status, language, and difficulty.}
\begin{tabular}{llr}
\toprule
\textbf{Category} & \textbf{Value} & \textbf{Number of Examples} \\
\midrule
Refusal Status & Refusal & 1013 \\
               & Non-refusal & 1232 \\
\midrule
Language       & English & 1048 \\
               & Korean & 1197 \\
\midrule
Difficulty     & SimpleQA & 743 \\
               & ComplexQA & 1502 \\
\bottomrule
\end{tabular}
\label{tab:dataset_stats}
\end{table}

\begin{table}[h]
\centering
\caption{Number of examples per reasoning type, grouped by difficulty type.}
\begin{tabular}{lrrr}
\toprule
\textbf{Reasoning Type} & \textbf{SimpleQA} & \textbf{ComplexQA} & \textbf{All} \\
\midrule
Multi-Constraint Reasoning & 357 & 1362 & 1719 \\
Numerical Reasoning        & 145 & 703  & 848  \\
Temporal Reasoning         & 99  & 700  & 799  \\
Tabular Reasoning          & 86  & 446  & 532  \\
Format Reasoning           & 55  & 368  & 423  \\
Textual Reasoning          & 0   & 170  & 170  \\
Others                     & 1   & 41   & 42   \\
\bottomrule
\end{tabular}
\label{tab:reasoning_types}
\end{table}

\newpage
\section{Prompts}
\label{sec:prompts}

\subsection{Prompts for Inference Methods}
\label{subsec:prompts-for-inference}

We employed various inference methodologies for diverse reasoning, and the prompts used are as follows: Direct Answer in \cref{fig:direct-answer-prompt}, CoT in \cref{fig:cot-prompt}, CoD in \cref{fig:cod-prompt}, Plan-and-Solve in \cref{fig:plan-and-solve-prompt}, and Least-to-Most in \cref{fig:least-to-most-prompt}.

\begin{figure}[htbp]
\centering
\begin{tcolorbox}[enhanced, colback=gray!5,colframe=green!70!black,title=Prompt for Direct Answer inference,
 coltitle=black, fonttitle=\bfseries, width=0.9\linewidth]

\textbf{User:} 
You are a helpful assistant tasked with answering questions strictly based on the content of the provided documents. The documents may contain irrelevant or inaccurate information, so please reason carefully and critically when forming your answer.
\\
<Rules>\\
1. Only use information that is explicitly stated in the documents. Do not rely on prior knowledge or make assumptions beyond the content.\\
2. If the question cannot be answered solely based on the provided documents, respond with:  
   "I cannot answer because the question is unanswerable with the documents."  
   Then briefly explain why the information is insufficient.\\
3. Always cite the document numbers used to derive your answer, using the format [1], [2], etc.\\
4. If multiple documents were referenced, include all relevant numbers at the end of your answer.\\
5. Answer the question directly. Do not return any preamble, explanation, or reasoning.\\
</Rules>\\

<Question>\\  
\{question\}  \\
</Question>  \\

<Documents>  \\
\{docs\}  \\
</Documents>\\

\end{tcolorbox}
\caption{Prompt used for Direct Answer inference in the Inference Methods experiment}
\label{fig:direct-answer-prompt}
\end{figure}

\begin{figure}[htbp]
\centering
\begin{tcolorbox}[enhanced, colback=gray!5,colframe=green!70!black,title=Prompt for CoT Inference,
 coltitle=black, fonttitle=\bfseries, width=0.9\linewidth]

\textbf{User:} 
You are a helpful assistant tasked with answering questions strictly based on the content of the provided documents. The documents may contain irrelevant or inaccurate information, so please reason carefully and critically when forming your answer.\\

<Rules>\\
1. Only use information that is explicitly stated in the documents. Do not rely on prior knowledge or make assumptions beyond the content.\\
2. If the question cannot be answered solely based on the provided documents, respond with:  
   "I cannot answer because the question is unanswerable with the documents."  
   Then briefly explain why the information is insufficient.\\
3. Always cite the document numbers used to derive your answer, using the format [1], [2], etc.\\
4. If multiple documents were referenced, include all relevant numbers at the end of your answer.\\
5. Think step by step to answer the following question. Return thinking steps between <Thinking> and </Thinking> and the answer between <Answer> and </Answer>.\\
</Rules>\\

<Question>\\  
\{question\}  \\
</Question>  \\

<Documents>  \\
\{docs\}  \\
</Documents> \\

\end{tcolorbox}
\caption{Prompt used for CoT inference in the Inference Methods experiment}
\label{fig:cot-prompt}
\end{figure}
\begin{figure}[htbp]
\centering
\begin{tcolorbox}[enhanced, colback=gray!5,colframe=green!70!black,title=Prompt for CoD Inference,
 coltitle=black, fonttitle=\bfseries, width=0.9\linewidth]

\textbf{User:} 
You are a helpful assistant tasked with answering questions strictly based on the content of the provided documents. The documents may contain irrelevant or inaccurate information, so please reason carefully and critically when forming your answer.\\

<Rules>\\
1. Only use information that is explicitly stated in the documents. Do not rely on prior knowledge or make assumptions beyond the content.\\
2. If the question cannot be answered solely based on the provided documents, respond with:  
   "I cannot answer because the question is unanswerable with the documents."  
   Then briefly explain why the information is insufficient.\\
3. Always cite the document numbers used to derive your answer, using the format [1], [2], etc.\\
4. If multiple documents were referenced, include all relevant numbers at the end of your answer.\\
5. Think step by step, but only keep a minimum draft for each thinking step, with 5 words at most. Return thinking steps between <Thinking> and </Thinking> and the answer between <Answer> and </Answer>.\\
</Rules>\\

<Question>  \\
\{question\}  \\
</Question>  \\
<Documents>  \\
\{docs\}  \\
</Documents>\\
\end{tcolorbox}
\caption{Prompt used for CoD inference in the Inference Methods experiment}
\label{fig:cod-prompt}
\end{figure}
\begin{figure}[htbp]
\centering
\begin{tcolorbox}[enhanced, colback=gray!5,colframe=green!70!black,title=Prompt for Plan-and-Solve Inference,
 coltitle=black, fonttitle=\bfseries, width=0.9\linewidth]

\textbf{User:} 
You are a helpful assistant tasked with answering questions strictly based on the content of the provided documents. The documents may contain irrelevant or inaccurate information, so please reason carefully and critically when forming your answer. \\

<Rules>\\
1. Only use information that is explicitly stated in the documents. Do not rely on prior knowledge or make assumptions beyond the content.\\
2. If the question cannot be answered solely based on the provided documents, respond with:  
   "I cannot answer because the question is unanswerable with the documents."  
   Then briefly explain why the information is insufficient.\\
3. Always cite the document numbers used to derive your answer, using the format [1], [2], etc.\\
4. If multiple documents were referenced, include all relevant numbers at the end of your answer.\\
5. First understand the problem and devise a plan to solve the problem. Then, carry out the plan to solve the problem step by step. Return intermidiate steps between <Thinking> and </Thinking> and the answer between <Answer> and </Answer>.\\
</Rules>\\

<Question>\\  
\{question\}  \\
</Question>  \\

<Documents>  \\
\{docs\}  \\
</Documents> \\

\end{tcolorbox}
\caption{Prompt used for Plan-and-Solve inference in the Inference Methods experiment}
\label{fig:plan-and-solve-prompt}
\end{figure}

\begin{figure}[htbp]
\centering
\begin{tcolorbox}[enhanced, colback=gray!5,colframe=green!70!black,title=Prompt for Least-to-Most inference,
 coltitle=black, fonttitle=\bfseries, width=0.9\linewidth]
\textbf{[Decomposition Prompt]}
\\
\textbf{User:} 
You are an assistant that decomposes a question into simpler sub-questions.

Context:\\
\{docs\}\\
Question: {question}\\
Return each sub-question on its own line, without numbering.\\

\textbf{[Solving Context Prompt]}
\\
\textbf{User:}
You are a helpful assistant tasked with answering questions strictly based on the content of the provided documents. The documents may contain irrelevant or inaccurate information, so please reason carefully and critically when forming your answer.\\

Rules:\\
1. Only use information that is explicitly stated in the documents. Do not rely on prior knowledge or make assumptions beyond the content.\\
2. If the question cannot be answered solely based on the provided documents, respond with:  
   I cannot answer because the question is unanswerable with the documents.  
   Then briefly explain why the information is insufficient.\\
3. Always cite the document numbers used to derive your answer, using the format [1], [2], etc.\\
4. If multiple documents were referenced, include all relevant numbers at the end of your answer.\\
5. Think step by step to answer the following question.\\

<Rules>\\
<Documents>\\
\{docs\}\\
</Documents>\\
\\
\textbf{[Solving Prompt]}
\\
\textbf{User:}
Answer the following question using the information provided in the context.\\
<Question>\\
\{sub\_question\}\\
</Question>
\end{tcolorbox}
\caption{Prompt used for Least-to-Most inference in the Inference Methods experiment}
\label{fig:least-to-most-prompt}
\end{figure}

\clearpage
\subsection{Prompts for QA Generation}
\label{subsec:prompts-for-qa-gen}

The QA generation process of our benchmark consists of four stages, with the prompts used for each as follows: KIE in \cref{fig:kie-prompt}, BasicQA in \cref{fig:basic-qa-prompt}, SimpleQA in \cref{fig:simple-qa-prompt}, and ComplexQA in \cref{fig:complex-qa-prompt}.

\begin{figure}[htbp]
\centering
\begin{tcolorbox}[
  enhanced,
  colback=gray!5,
  colframe=green!70!black,
  title=KIE Generation Prompt,
  coltitle=black,
  fonttitle=\bfseries,
  width=0.9\linewidth,
]
\textbf{User:}
You will be provided with \{chunkcount\} chunk(s) of document, each in \{format\} format.\\
Your mission is to come up with a series of questions with varying levels of difficulty to test students' understanding of the information contained within these document chunks.\\
This process will be conducted over multiple steps.\\
Use \{language\} language to generate the key, value, and description.\\

Here are the document chunks:\\
\{content\}\\
{[First task]}\\
Your first task is to generate a collection of Key-Value pairs for a Key-Information Extraction (KIE) task.\\
The extracted key information should have sufficient coverage and be comprehensive enough to reconstruct a text passage that retains the essential ideas and meaning of the original text.\\
Please provide the Key-Value pairs in the following format:\\
\textasciigrave \textasciigrave \textasciigrave json\\
\{\{\\
    "KIE": [\\
        \{\{\\
            "Key": "<Name of key 1>",\\
            "Value": "[List of values corresponding to the key found in the text chunks]",\\
            "Description": "<Description of the key to support understanding of the Key-Value pair>"\\
        \}\},\\
        \{\{\\
            "Key": "<Name of key 2>",\\
            "Value": "[List of values corresponding to the key found in the text chunks]",\\
            "Description": "<Description of the key to support understanding of the Key-Value pair>"\\
        \}\},\\
        ...\\
    ]\\
\}\}\\
\textasciigrave \textasciigrave \textasciigrave \\

You are to only respond in JSON format and providing the Key-Value pairs for all the chunks supplied.
\end{tcolorbox}

\caption{Prompt used for KIE Generation stage.}
\label{fig:kie-prompt}
\end{figure}

\begin{figure}[htbp]
\centering
\begin{tcolorbox}[
  enhanced,
  colback=gray!5,
  colframe=green!70!black,
  title=Basic QA Generation Prompt,
  coltitle=black,
  fonttitle=\bfseries,
  width=0.9\linewidth,
]
\textbf{User:}
{[Second task]}\\
Your task is to create a set of simple Q\&A pairs using the key-value pairs extracted from the document, while also referring to the original document chunks.\\
These Q\&A pairs should align with the following characteristics:\\
    1.	Question Construction: Each question should focus on the key from the key-value pairs, while the corresponding value provides the answer.\\
	2.	Contextual Independence: The questions should not explicitly reference the document or assume its availability. This is because the students answering these questions are expected to have internalized the document’s content. Phrases like “In the document…” should be avoided.\\
    3.	Reasoning Type: Each question should test one of the following reasoning types:\\
	    •	Form/Layout Understanding: The question assesses the student’s ability to comprehend the document’s layout or structure, rather than just its text content.\\
	    •   Tabular Understanding: The question evaluates the student’s ability to interpret and extract information from tabular data.\\
	    •	Text/Semantic Understanding: The question examines the student’s grasp of the textual content’s meaning and implications.\\
Use \{language\} language to generate the questions and answers.\\

The Q\&A pairs should be in the following format:\\
\textasciigrave \textasciigrave \textasciigrave json\\
\{\{\\
    "simpleQA": [\\
        \{\{\\
            "id": "<ID of the simpleQA, e.g. simpleQA1, simpleQA2, ...>",\\
            "question": "<Question text>",\\
            "answer": "<Answer text>",\\
            "reasoning\_type": "[List of reasoning types that the question tests, e.g. Form/Layout Understanding, Tabular Understanding, Text/Semantic Understanding]"\\
        \}\},\\
        ...\\
    ]\\
\}\}\\
\textasciigrave \textasciigrave \textasciigrave \\

You are to respond in JSON format only and ensure the questions are clear, concise, and directly tied to the key-value pairs provided.
\end{tcolorbox}

\caption{Prompt used for Basic QA Generation stage.}
\label{fig:basic-qa-prompt}
\end{figure}

\begin{figure}[htbp]
\centering
\begin{tcolorbox}[
  enhanced,
  colback=gray!5,
  colframe=green!70!black,
  title=Simple QA Generation Prompt,
  coltitle=black,
  fonttitle=\bfseries,
  width=0.9\linewidth,
]
\textbf{User:}
[Third task]\\
Your next task is to refer to the simple Q\&A pairs created in the previous turn and generate a new set of Q\&A pairs that require application reasoning.\\
Application reasoning refers to Q\&As that require additional steps or though processes to answer, as opposed to simple query-and-fetch Q\&A.\\
Use \{language\} language to generate the questions and answers.\\
Each question should be clear, consider and aligned with the characteristics and reasoning types described below:\\

Reasoning Types:\\
1. Numerical reasoning: The question requires the reader to perform arithmetic operations on the information provided in the document, such as counting, comparisons, calculations, etc.\\
2. Tabular reasoning: The question requires the reader to compare and contrast information across different tables, rows, columns, etc.\\
3. Multi-constraint reasoning: The question which contains multiple conditions / constraints which require readers to find the answer that satisfies all the conditions / constraints.\\
4. Temporal reasoning: The question requires the reader to reason about the time-based information provided in the document.\\
5. Format reasoning: The question requires the reader to reason about the format / post-processing of the information provided in the document (e.g. conversion of units, etc.).\\

Modify existing or refer to the simple Q\&A pairs, or add new ones to incorporate application reasoning types.\\
Ensure each Q\&A is self-contained and does not explicitly reference the document (e.g., avoid phrases like “In the document…”).\\
You are to response in the following JSON format:\\
\textasciigrave \textasciigrave \textasciigrave json\\
\{\\
    "simpleAppQA": [\\
        \{\\
            "id": "<ID of the simpleAppQA, e.g. simpleAppQA1, simpleAppQA2, ...>",\\
            "question": "<Question text>",\\
            "answer": "<Answer text>",\\
            "reasoning\_type": "<Reasoning type of the question, e.g. Numerical reasoning, Tabular reasoning, Multi-constraint reasoning, Temporal reasoning, Format reasoning>"\\
        \},\\
        ...\\
    ]\\
\}\\
\textasciigrave \textasciigrave \textasciigrave \\

You are to respond in JSON format only and ensure the questions are clear, concise, and require reasoning beyond simple query-and-fetch.
\end{tcolorbox}

\caption{Prompt used for Simple QA Generation stage.}
\label{fig:simple-qa-prompt}
\end{figure}

\begin{figure}[htbp]
\centering
\begin{tcolorbox}[
enhanced,
colback=gray!5,
colframe=green!70!black,
title=Complex QA Generation Prompt,
  coltitle=black,
  fonttitle=\bfseries,
  width=0.9\linewidth,
  ]
\textbf{User:}
[Final task]\\
Now, we will create complex application QAs based on the existing simple application QAs.\\
These questions differ from simple application QAs by requiring multiple reasoning steps and incorporating a combination of reasoning types to arrive at the answer.\\
Use \{language\} language to generate the questions and answers.\\

When forming complex/multiple application QAs, follow these guidelines:\\
	1.	Merge and Modify Thoughtfully:\\
        •	Combine information from different simple application QAs to form new, complex questions.\\
        •	Avoid creating trivial questions that are merely a concatenation of existing QAs. Ensure the merged question requires deeper reasoning and processing.\\
	2.	Step-by-Step Reasoning:\\
        •	Frame the question so that the student must:\\
            •	Utilize the reasoning result of one step as an input for the next step.\\
            •	Apply additional reasoning (numerical, temporal, tabular, multi-constraint, or format reasoning) to arrive at the final answer.\\
	3.	Challenge and Engagement:\\
        •	Ensure the QAs challenge the student by requiring them to integrate knowledge and think critically.\\
        •	Design the reasoning flow to be logical and non-trivial.\\

The generated complex application QAs should be in the following JSON format:\\
\textasciigrave\textasciigrave\textasciigrave json\\
\{\\
    "complexQA": [\\
        \{\\
            "id": "<ID of the complexQA, e.g. complexQA1, complexQA2, ...>",\\
            "question": "<Question text>",\\
            "answer": "<Answer text>",\\
            "reasoning\_type": "[List of reasoning types that the question tests, e.g. Numerical reasoning, Tabular reasoning, Multi-constraint reasoning, Temporal reasoning, Format \\reasoning]"\\
        \},\\
        ...\\
    ]\\
\}\\
\textasciigrave\textasciigrave\textasciigrave\\

You are to respond in JSON format only and ensure the questions are clear, concise, and require multiple reasoning steps to arrive at the answer.\\
Generating these QAs could be challenging, but it will help students develop a deeper understanding of the content.
\end{tcolorbox}

\caption{Prompt used for Complex QA Generation stage.}
\label{fig:complex-qa-prompt}
\end{figure}

\clearpage
\subsection{Prompt for LLM Evaluation}
\label{subsec:prompt-for-llm-eval}

Our benchmark, CReSt, utilizes LLM Evaluation for assessment in non-refusal environments, with the prompt used for this shown in \cref{fig:rating-prompt}.

\begin{figure}[htbp]
\centering
\begin{tcolorbox}[enhanced, colback=gray!5,colframe=green!70!black,title=Rating Prompt,
  coltitle=black, fonttitle=\bfseries, width=0.9\linewidth]

\textbf{User:} 
You are an evaluator for a Retrieval Question Answering (QA) task. Your task is to assess how closely the predicted answer matches the golden answer.\\
\\
**Evaluation Categories:**\\
- **Correct**: The predicted answer is a perfect match or semantically identical to the golden answer.\\
- **Partially Correct**: The predicted answer contains some key information from the golden answer but may be incomplete, missing details, or only partially aligned.\\
- **Wrong**: The predicted answer is completely incorrect, missing essential details, or contains misleading information.\\

**Consider the following factors when evaluating:**\\
- **Exactness**: Does the predicted answer exactly match the golden answer?\\
- **Paraphrasing**: If reworded, does it retain the same meaning?\\
- **Completeness**: Is the full answer provided, or is it partial?\\
- **Incorrect Information**: Does the predicted answer introduce any false or misleading details?\\

**Error Category Guidelines:**\\
*If the evaluation is not **Correct** (i.e., it is either "Partially Correct" or "Wrong"), also identify the most severe error type present by providing an **ErrorType** field. This\\ field should contain one of the following categories that best describes the main error:*\\

- **AnswerRefusal**: The answer refuses to provide a response or gives up on answering, despite a clear expectation to do so.\\
- **NumericMistakes**: The answer contains incorrect arithmetic or inaccurate numeric references (e.g., population sizes, years, differences in ages, championship tallies).\\
- **MissingDetail**: The answer shows a partial understanding by identifying the correct domain or background but omitting the necessary numeric or textual detail.\\
- **Others**: Any other error types not covered by the above categories.\\

**Input:**\\
- **Question**: \{question\}\\
- **Golden Answer**: \{golden\_answer\}\\
- **Predicted Answer**: \{predicted\_answer\}\\

**Your response should be formatted as follows:**\\

\textasciigrave\textasciigrave\textasciigrave plaintext\\
**Justification**: <brief explanation of your evaluation>\\
**Decision**: <Correct/Partially Correct/Wrong>\\
**ErrorType**: <AnswerRefusal/NumericMistakes/MissingDetail/Others>  // Keep empty if the evaluation is **Correct**\\
\textasciigrave\textasciigrave\textasciigrave

\end{tcolorbox}
\caption{Prompt used to automatically rate the answers of LLMs in the experiments.
}
\label{fig:rating-prompt}
\end{figure}

\end{document}